%% file: main.tex
\documentclass[onecolumn,english,twocolumn,conference]{IEEEtran}

\usepackage[T1]{fontenc}
\usepackage[utf8]{inputenc}
\usepackage{babel}
\usepackage{array}
\usepackage{float}
\usepackage{booktabs}
\usepackage{mathtools}
\usepackage{multirow}
\usepackage{amstext}
\PassOptionsToPackage{normalem}{ulem}
\usepackage{ulem}
\usepackage[]
 {hyperref}

\makeatletter

\providecommand{\tabularnewline}{\\}
\floatstyle{ruled}
\newfloat{algorithm}{tbp}{loa}
\providecommand{\algorithmname}{Algorithm}
\floatname{algorithm}{\protect\algorithmname}

\usepackage{cite}
\usepackage{amsmath,amssymb,amsfonts}
\usepackage{graphicx}
\usepackage{textcomp}
\usepackage{xcolor}
\usepackage{algpseudocode}

\makeatother

\begin{document}
\title{QuAKE: Speeding up Model Inference Using \uline{Qu}ick and \uline{A}pproximate
\uline{K}ernels for \uline{E}xponential Non-Linearities}
\author{\IEEEauthorblockN{Sai Kiran Narayanaswami} \IEEEauthorblockA{{\small\emph{Centre for Responsible AI, }}\\
{\small\emph{Indian Institute of Technology, Madras}}{\small}\\
{\small\texttt{saikirann94@gmail.com}}} \and \IEEEauthorblockN{Gopalakrishnan Srinivasan} \IEEEauthorblockA{{\small\emph{Robert Bosch Centre for Data Science and AI}}{\small ,}{\small\emph{}}\\
{\small\emph{Dept. of Computer Science and Engineering,}}\\
{\small\emph{Indian Institute of Technology, Madras}}{\small{} }\\
{\small\texttt{sgopal@cse.iitm.ac.in}}} \and \IEEEauthorblockN{Balaraman Ravindran} \IEEEauthorblockA{{\small\emph{Wadhwani School of Data Science and AI, }}\\
{\small\emph{Indian Institute of Technology, Madras}}{\small}\\
{\small\texttt{ravi@dsai.iitm.ac.in}}}}
\maketitle
\begin{abstract}
As machine learning gets deployed more and more widely, and model
sizes continue to grow, improving computational efficiency during
model inference has become a key challenge. In many commonly used
model architectures, including Transformers, a significant portion
of the inference computation is comprised of exponential non-linearities
such as Softmax. In this work, we develop QuAKE, a collection of novel
operators that leverage certain properties of IEEE-754 floating point
representations to quickly approximate the exponential function without
requiring specialized hardware, extra memory, or precomputation. We
propose optimizations that enhance the efficiency of QuAKE in commonly
used exponential non-linearities such as Softmax, GELU, and the Logistic
function. Our benchmarks demonstrate substantial inference speed improvements
between 10\% and 35\% on server CPUs, and 5\% and 45\% on embedded
and mobile-scale CPUs for a variety of model architectures and sizes.
Evaluations of model performance on standard datasets and tasks from
various domains show that QuAKE operators are able to provide sizable
speed benefits with little to no loss of performance on downstream
tasks.
\end{abstract}


\input{introduction.tex}

\input{background.tex}

\input{methods.tex}

\input{experiments.tex}

\input{related-work.tex}

\input{endmatter.tex}

\end{document}

%% file: introduction.tex
\section{Introduction}

Machine Learning (ML) has become pervasive, with recent breakthroughs
in large-scale ML technologies such as ChatGPT\cite{chatgpt} only
serving to accelerate the deployment of ML models. Such progress,
however, has come at the cost of burgeoning resource usage (during
both training and deployment) as models continue to grow in size.
Thus, there are strong socioeconomic and ecological incentives to
reduce the resource footprint of ML in general. In this work, we focus
on inference latency, a key factor in determining the post-training
footprint of a model. Improving the latency of inference algorithms
provides the following two-fold benefits. First, doing so enhances
the effectiveness of model deployments. For instance, lower inference
latency leads to faster responses in real-time applications such as
Automatic Speech Recognition (ASR). More importantly, lower latency
often leads to correspondingly lower energy consumption, while also
expanding the scope for deploying models on less powerful hardware.

Non-linear activations are crucial for the effectiveness of neural
networks in general\cite{univ-approx-thm}. Exponential-based layer
activations such as GELU, which is widely used in contemporary models,
have been shown to improve model performance compared to simpler ones
such as ReLU\cite{gelu}. The Softmax operation is especially critical
to the now-ubiquitous Transformer\cite{transformers} architecture
used by the vast majority of Large Language Models (LLMs), as well
as many other kinds of models. Although effective, non-linearities
of this class are computationally intensive. As our measurements in
Table \ref{table:models} show, layer activations can account for
as much as a quarter or more of the inference time in some models,
while softmax can account for 10\% or more in others.

In this work, we focus on techniques to improve the speed of these
operations, primarily by speeding up the computation of the exponential
function. Approaches proposed by existing works involve approximating
the exponential using lookup tables\cite{nn-lut} or replacing it
with computationally simpler operations\cite{peano-vit,ml-plac},
possibly by implementing specialized hardware architectures\cite{hw-efficient-softmax,softermax}.
Each comes with its own trade-offs between speed, memory and bandwidth
usage, and model performance.

Our work expands on the technique proposed by Schraudolph\cite{schraudolph-exp}
that makes use of certain properties of IEEE-754 floating point representations\cite{ieee754-2019}
to quickly approximate the exponential. We propose a novel adaptation
that enables affine input transformations at zero additional cost,
leading to significant savings in computation. To improve accuracy,
we also develop second-order refinements of these operators, similar
to the extensions by \cite{exp-mantissa-poly}. We use the label \textbf{QuAKE}
(\textbf{Qu}ick and \textbf{A}pproximate \textbf{K}ernels for \textbf{E}xponentiation)
to collectively refer to our proposed ideas, as well as in the names
of the individual operators defined herein.

Unlike most prior works, QuAKE requires no specialized hardware, extra
memory, or precomputation. Consisting solely of algorithmic changes
to the computation of non-linearities, QuAKE is not restricted to
specific classes of models. Further enhancing the applicability of
QuAKE are factors like range insensitivity (i.e. maintaining approximation
accuracy over the entire numerical range) and amenability to vectorization.

We conduct extensive CPU benchmarks to measure the speedups in model
inference due to QuAKE operators by implementing them in the TensorFlow
Lite\cite{tensorflow2015-whitepaper} inference framework. For these
benchmarks, we use a variety of standard model architectures (Transformers
and Convolutional networks) and sizes, as well as various types of
CPUs from Single Board Computers (SBC) such as the \href{https://www.raspberrypi.com/products/}{Raspberry Pi}
series and the VisionFive 2\cite{vf2-datasheet} RISC-V board, to
server CPUs. QuAKE operators provide up to 30\% speedup in inference,
even when using the more accurate second-order versions. 

To quantify the effects of approximation on output quality, we compare
model performance with and without QuAKE on several standard datasets.
These datasets involve downstream tasks such as image classification
(Imagenet ILSVRC 2012\cite{imagenet-data}), ASR (LibriSpeech\cite{librispeech})
and language understanding/generation (Hellaswag\cite{hellaswag}).
We find that the use of QuAKE operators, even with the less accurate
first-order forms, does not degrade model performance considerably.
In fact, the second-order versions lead to almost no loss in task
performance compared to reference implementations.

To the best of our knowledge, we are the first to specifically study
this class of methods in the context of model inference. Our results,
along with the many desirable properties we highlight, serve to establish
QuAKE operators as ideal drop-in replacements for exponential non-linearities
in model inerence frameworks.

%% file: background.tex
\section{Background}

\subsection{IEEE-754 Floating Point Representations}\label{subsec:IEEE-754-Floating-Point}

The IEEE-754 formats\cite{ieee754-2019} represent floating point
numbers using 3 bit fields: the sign bit, exponent bits, and the mantissa
bits. The formats are defined for various precisions, most notably
the 32-bit \emph{single precision} format. We use this format throughout,
although our ideas are just as easily applicable to other precisions.

In general, there are $l_{e}$ exponent bits, followed by $l_{m}$
mantissa bits. The exponent bits are biased by a value $B$, known
as the \emph{exponent bias}. That is, an exponent value $x_{e}$ is
specified by a biased value of $x_{e}+B$, written as an unsigned
integer of $l_{e}$ bits. The mantissa bits indicate a fractional
value $x_{m}\in[0,1)$ whose bits after the binary point are given
by those in the mantissa field. The magnitude of the number represented
by these bits is given by
\begin{equation}
|x|=2^{x_{e}}\left(1+x_{m}\right)\label{eq:fp-main}
\end{equation}
where $|.|$ denotes the magnitude operator. The sign bit is zero
if the number is non-negative, and one otherwise. In the single precision
format, $l_{e}=8$, $l_{m}=23$, and $B=127$.

\subsection{Exponentiation via Floating Point Representations}\label{subsec:Exponentiation-via-Floating}

Here, we introduce the general principle involved in utilizing the
structure of floating point representations to quickly approximate
exponentials. A real number $x$ can be expressed as the sum of its
integral part (or floor), $\left\lfloor x\right\rfloor $, and its
fractional part, $\{x\}$. Consequently, its exponential (to the base
2) can be written as
\begin{equation}
2^{x}=2^{\left\lfloor x\right\rfloor }2^{\{x\}}\approx2^{\left\lfloor x\right\rfloor }\left(1+\{x\}\right)\label{eq:quake-result}
\end{equation}
where we have approximated $2^{\{x\}}$ by the secant line $1+\{x\}$
in the interval $[0,1)$, which is the range of $\{x\}$. This approximation
corresponds to a floating point value whose exponent field is set
to $\left\lfloor x\right\rfloor +B$ and its mantissa to the binary
representation of $\{x\}$ after the decimal point, as shown in Equation
\ref{eq:fp-main}. Therefore, computing this approximate exponential
is equivalent to assembling the bit-representation of such a number.
Consider the bit-representation of the approximate result viewed as
an integer. The \emph{value} of this integer, $z$, is given by
\begin{equation}
z=\left(\left\lfloor x\right\rfloor +B\right)\cdot2^{l_{m}}+\{x\}\cdot2^{l_{m}}=\left(x+B\right)\cdot2^{l_{m}}\label{eq:quake-derivation}
\end{equation}

Thus, computing $z$ amounts to an ordinary affine transformation
of the input $x$. $z$ can be converted by value from floating point
form to an integer to obtain the requisite bit representation, which
can then be reinterpreted as a floating point value in high-level
code (e.g. by using the \texttt{\textbf{reinterpret\_cast}} keyword
in C++) at no additional computational cost. In our pseudocode, we
refer to this reinterpretation as \texttt{\textbf{int\_bits\_as\_float}},
while \texttt{\textbf{convert\_to\_int}} refers to value conversion
from floating point to integer.

\subsection{GELU Activation Function}

The Gaussian Error Linear Unit (GELU)\cite{gelu} is given by $\text{GELU}\left(x\right)=x\Phi\left(x\right)$,
where $\Phi(x)$ is the Cumulative Distribution Function of the standard
Gaussian distribution. Due to the high computational cost of this
function, it is often replaced with the following approximation when
suitable implementations of $\Phi(x)$ are not available:
\begin{equation}
\text{GELU}\left(x\right)\approx0.5x\left(1+\tanh\left[\sqrt{2/\pi}\left(x+0.044715x^{3}\right)\right]\right)\label{eq:gelu-tanh}
\end{equation}

%% file: methods.tex
\section{Proposed Methods}\label{sec:Methods}

\subsection{QuAKE}\label{subsec:QuAKE}

The procedure described in Section \ref{subsec:Exponentiation-via-Floating}
approximates $2^{x}$ for a given input $x$. To compute the natural
exponential, $e^{x}$, one could perform the same steps with the input
$x\log_{2}e$ instead. The additional multiplication can be avoided
by pre-multiplying $\log_{2}e$ with $2^{l_{m}}$ in Equation \ref{eq:quake-derivation}.
We observe, more generally, that any affine transform $px+q$ with
known coefficients $p$ and $q$ can be incorporated into Equation
\ref{eq:quake-derivation}, leading to the general form
\begin{equation}
z=c_{0}x+c_{1},\;c_{0}=2^{l_{m}}p,\ c_{1}=2^{l_{m}}(B+q)\label{eq:quake-affine}
\end{equation}

The above observation motivates the formulation of a family of functions
parameterized by $c_{0}$ and $c_{1}$, which we term \textbf{QuAKE}
(\textbf{Qu}ick and \textbf{A}pproximate \textbf{K}ernels for \textbf{E}xponentiation).
For instance, to calculate the standard exponential in single-precision
($B=127$), one would use $c_{0}$$=$$2^{23}\log_{2}e$ and $c_{1}=127\cdot2^{23}$.
Following \cite{schraudolph-exp}, we also add a bias value to $c_{1}$
that minimizes the maximum relative approximation error, giving $c_{1}=(127-0.0436)\cdot2^{23}$.

\subsection{QuAKE for Improved Approximation Accuracy}\label{subsec:QuAKE2}

While the approximation of $2^{\{x\}}$ in Equation \ref{eq:quake-result}
proves to be reasonably accurate for use in model inference, as shown
in Section \ref{subsec:Downstream-Task-Performance}, it still has
a maximum relative approximation error of about 4.3\%\cite{schraudolph-exp}.

To reduce this error, we derive a second-order approximation to $2^{\{x\}}$
that can be efficiently evaluated. One could consider a best-fit quadratic
function of $\{x\}$ that minimizes the average or maximum error in
the interval $[0,1)$. Since the result $z$ of QuAKE (base-2) already
has $\{x\}$ as the mantissa, setting its exponent field to $B$ (i.e.
a biased exponent of zero) would yield the value $a_{m}:=1+\{x\}$.
This operation is easily performed using two bit-wise logical instructions
to first mask out the original exponent, and then set it to $B$.
The quadratic approximation to $2^{\{x\}}$ may be equivalently rewritten
as
\begin{equation}
2^{\{x\}}\approx(a_{0}a_{m}{}^{2}+a_{1}a_{m}+a_{2})\coloneqq y_{m}\label{eq:quake2-poly}
\end{equation}

An exhaustive grid search for the set of coefficients that minimize
the worst case relative error of the approximation over the interval
$[0,1)$ yielded coefficients $a_{0}=0.33$, $a_{1}=-0.017$, and
$a_{2}=0.68$, with an empirically estimated maximum relative error
of 0.17\%.

Continuous approximations are desirable in model inference, as continuity
prevents sudden jumps in the output due to input noise. It is easily
seen that continuity is preserved as long as our approximation for
$2^{\{x\}}$ is accurate for $\{x\}=0$ and in the limit of $\{x\}\rightarrow1$,
i.e. $y_{m}=$ 1 and 2 when $a_{m}=$1 and 2 respectively. We set
$a_{1}=0$ to simplify the computation further. These conditions imply
$a_{0}=1/3$, and $a_{2}=2/3$. We empirically estimated the maximum
deviation of this approximation over $[0,1)$ to be 0.34\%. Although
marginally less accurate than the optimum found earlier, we found
no apparent difference in model performance between the two sets of
parameters. Being simple rational numbers, these coefficients are
also more suited for hardware implementations. Therefore, we make
use of these values in subsequent discussion and experiments.

Once this corrected mantissa (plus one) is obtained, the final result
$y$ can be obtained by simply joining the exponent field of the QuAKE
result $z$ and the mantissa field of $y_{m}$, followed by reinterpretation
as a floating point value. We refer to this refined approximation
procedure as QuAKE2, and describe it in Algorithm \ref{algo:quake2}.
Although QuAKE2 adds several additional steps to QuAKE, most of them
involve light bit-wise operations. Its computational cost is, therefore,
substantially lower than the higher-order polynomial approximations
in standard implementations. Indeed, as we show in Section \ref{subsec:Speedup-affine},
QuAKE2 yields massive operator speedups comparable to those from QuAKE.
Note that just like QuAKE, affine input transformations can be achieved
at no cost with QuAKE2. Only $c_{0}$ and $c_{1}$ need to be set
as in Equation \ref{eq:quake-affine}, and the coefficients from Equation
\ref{eq:quake2-poly} need not be changed.

\begin{algorithm}
\sf
\begin{algorithmic}[1]
\Function{QuAKE2}{

$x$, $c_{0}$, $c_{1}$}

\State{\texttt{int} $M_{e,0}:=$ \texttt{0x007FFFFF;}}

\State{\texttt{int} $M_{m,0}:=$ \texttt{0xFF800000;}}

\State{\texttt{int} $M_{e,127}:=$ \texttt{0x3F800000;}}

\State{\texttt{float} $l:=\ x\cdot c_{0}+c_{1}$;}

\State{\texttt{int} $z:=$ \texttt{\textbf{convert\_to\_int}}\texttt{($l$);}}

\State{\texttt{int} $m:=$ $(z\ \mathtt{\&}\ M_{e,0})\ \mathtt{|}\ M_{e,127})$\texttt{;}}

\State{\texttt{float} $a_{m}:=$ \texttt{\textbf{int\_bits\_as\_float}}\texttt{($m$);}}

\State{$a_{m}:=\ (a_{m}^{2}+2)\ /\ 3$\texttt{;}}

\State{\texttt{int} $a_{m,i}:=$ \texttt{\textbf{float\_bits\_as\_int}}\texttt{($a_{m}$);}}

\State{\texttt{int} $y_{i}:=$ $(z\ \mathtt{\&}\ M_{m,0})\ +\ a_{m,i}-(127\cdot2^{23}))$\texttt{;}}

\State{\texttt{float} $y:=$ \texttt{\textbf{int\_bits\_as\_float}}\texttt{($y_{i}$);}}

\State{\Return $y$\texttt{;}}

\EndFunction
\end{algorithmic}

\caption{QuAKE2}
\label{algo:quake2}
\end{algorithm}

\subsection{SoftMax}

SoftMax operations involve scaling each component of an input vector,
$v$, by a constant scale (or \emph{temperature}) parameter, $t$.
Additionally, for numerical stability, it is common practice to subtract
from each input value $v_{i},\ i=1\ldots N$ the maximum value in
$v$ to prevent overflow during exponentiation. Although this maximum
value is only known at runtime, it remains unchanged after it is computed.
Effectively, the scaling and maximum subtraction steps constitute
a fixed affine transformation that can be fused into QuAKE (or QuAKE2)
as described in Section \ref{subsec:QuAKE}. The resulting algorithm
is described in Algorithm \ref{algo:softmax}.

\begin{algorithm}
\sf
\begin{algorithmic}[1]
\Function{SoftMax-QuAKEx}{

$v$, $t$, $N$, $y$}

\State{\texttt{float $m:=\max_{i=1\ldots N}v_{i}$;}}

\State{\texttt{float $c_{0}:=\ 2^{23}\cdot t/\ln2$;}}

\State{\texttt{float $c_{1}:=\ 127\cdot2^{23}-c_{0}m$;}}

\State{\texttt{float $v_{min}:=\ \mathtt{MIN\_EXPONENT}+mt/\ln2$;}}

\State{\texttt{float $\mathrm{sum}:=\ 0$;}}

\For{$i=1,\ 2\ldots,\ N$}

\State{\texttt{$y_{i}:=$}\textsc{QuAKEx}\texttt{$\left(\max(v_{i},\ v_{min}),\ c_{0},\ c_{1}\right)$;}}

\State{\texttt{$\mathrm{sum}:=\mathrm{sum}+y_{i}$;}}

\EndFor

\For{$i=1,\ 2\ldots,\ N$}

\State{\texttt{$y_{i}:=y_{i}/\mathrm{sum}$;}}

\EndFor

\EndFunction
\end{algorithmic}

\caption{SoftMax with QuAKE or QuAKE2}
\label{algo:softmax}
\end{algorithm}

\subsection{GELU}

It is easy to see that Equation \ref{eq:gelu-tanh} for GELU is equivalent
to
\begin{equation}
\text{GELU}\left(x\right)\approx x\cdot\sigma\left(2\cdot0.044715\sqrt{\frac{2}{\pi}}x\left(\frac{1}{0.044715}+x^{2}\right)\right)\label{eq:gelu-quake2}
\end{equation}
where $\sigma(x)=(1+e^{-x})^{-1}$ is the standard logistic (sigmoid)
function. This re-expression allows more multiplicative constants
to be fused into QuAKE (or QuAKE2), as well as efficient evaluation
of the term $\frac{1}{0.044715}+x^{2}$ using a single Fused Multiply-Add
(FMA) instruction where supported.

%% file: experiments.tex
\section{Experimental Setup}

\subsection{Hardware}\label{subsec:Hardware}

For speed benchmarks, we use 4 different pieces of hardware, ranging
from embedded to server scales, to demonstrate the efficacy of the
proposed QuAKE operators. Their specifications are summarized in Table
\ref{table:hardware}.

\begin{table}
\centering
\begin{tabular}[b]{cccc}
\toprule 
Hardware & Clock Speed & Memory & Cache (L1/L2/L3)\tabularnewline
\midrule 
Raspberry Pi 5\cite{rp5-datasheet} & 1.5 GHz & 4 GB & 0.25/2/2 MiB\tabularnewline
\begin{minipage}[c][2.5em]{7em}%
\begin{center}
Raspberry Pi Zero 2W\cite{rp0-datasheet}
\par\end{center}%
\end{minipage} & 1.0 GHz & 512 MB & 32/512/- KiB\tabularnewline
Vision-Five 2\cite{vf2-datasheet} & 1.5 GHz & 8 GB & 32K/2MiB/-\tabularnewline
AMD EPYC 7282\cite{epyc-7282} & 2.79 GHz & 86 GB & 1/8/64 MiB\tabularnewline
\end{tabular}\caption{Specifications for hardware used. Note that effective clock speed
is reported, not maximum.}
\vspace{-2.5em}
\label{table:hardware}
\end{table}

\subsection{Models}\label{subsec:Models}

We use the following models in our experiments:

\paragraph*{Vision Transformers}

ViTs\cite{vit-og} use the transformer architecture to process images.
Various model sizes and configurations, labeled Tiny (Ti), Base (B),
and Large (L), are used for the benchmarking and performance studies
reported in this work. In addition, we include Swin\cite{swin} (Tiny),
which is a backbone model that uses a hierarchical transformer design.

\paragraph*{Convolutional Neural Networks (CNN)}

ConvNeXt-Tiny\cite{convnext} was proposed as a purely convolutional
alternative to ViTs that could achieve performance comparable to ViTs.
We also benchmark the YOLO-v8\cite{yolov8-website} and the FastSam-S\cite{fastsam}
semantic segmentation models. Both are CNN models that use the logistic
function.

\paragraph*{Whisper}

These are a series of ASR transformer models\cite{whisper} with an
encoder-decoder architecture. We use the Tiny, Small, and Base configurations
of these models, trained for English language transcription.

\paragraph*{LLMs}

The OPT (Open Pre-trained Transformers)\cite{opt} models are publicly
available LLMs based on GPT-3\cite{gpt3}. We use an OPT model with
1.3B parameters, along with GPT2-Large and GPT2-XL\cite{gpt2}.

The models are listed in Table \ref{table:models} along with relevant
details such as the activations involved. For Whisper models, the
parameter count is reported for encoder and decoder portions respectively.
The model weights were sourced from \href{https://huggingface.co/models}{Huggingface},
\href{https://aihub.qualcomm.com/}{Qualcomm AI Hub}, or their respective
publications.

\begin{table}
\begin{centering}
\begin{tabular}{cccc}
\toprule 
Model & Params & Softmax(\%) & GELU(\%)\tabularnewline
\midrule 
YOLO-v8  & 3.4M & 0.2  & 9.6$^{\dag}$\tabularnewline
ViT-Ti  & 8.2M & 8.9  & 17.1 \tabularnewline
Whisper-Ti(E/D)  & 9.4/28.0M & 12.2/1.0  & 11.2/0.7\tabularnewline
FastSAM-S  & 11.8M & 0.1  & 5.6$^{\dag}$\tabularnewline
Whisper-B(E/D)  & 23.7/48.6M & 10.6/1.1 & 9.5/0.8 \tabularnewline
Swin-Ti  & 29M & 2.1  & 18.3 \tabularnewline
ViT-B8  & 50M & 5.6  & 8.5 \tabularnewline
ConvNext-Ti  & 60M & - & 25.0 \tabularnewline
Whisper-S(E/D)  & 102/139M & 8.3/1.4 & 7.0/1.0 \tabularnewline
ViT-L  & 307M & 3.4  & 7.4 \tabularnewline
GPT2-L  & 774M & 6.1  & 4.4 \tabularnewline
OPT-1.3B  & 1.3B & 6.6  & -\tabularnewline
GPT2-XL  & 1.5B & 5.3  & 4.0 \tabularnewline
\midrule
\multicolumn{4}{l}{$^{\dag}$Activation is the Logistic function}\tabularnewline
\end{tabular}
\par\end{centering}
\caption{Models used in the experiments and percentages of their inference
runtimes accounted for by non-linearities, measured on the AMD EPYC
7282.}
\vspace{-1.5em}
\label{table:models}

\end{table}

\section{Results}

The experiments we present here aim to empirically answer the following
questions:
\begin{enumerate}
\item How much speedup do QuAKE operators provide for model inference with
standard model architectures?
\item How much speedup do the optimizations proposed in Section \ref{sec:Methods}
provide over simply replacing exponential function calls with QuAKE?
\item How does the use of the approximate operators affect model performance
on downstream tasks across application domains, and what tradeoffs
exist, if any, between inference speed and task performance?
\item How do model and hardware characteristics, especially scale, affect
the results?
\end{enumerate}

\subsection{Inference Speed}\label{subsec:Inference-Speed}

To benchmark inference speed for the models described in Section \ref{subsec:Models},
we use TensorFlow Lite (TFLite)\cite{tensorflow2015-whitepaper},
which we build with QuAKE operators. Benchmarking is done using the
built-in benchmark tool in TFLite. Model inference is run on randomized
inputs for 100 iterations, and up to 1 hour per model, and the average
inference time is used to measure speedups over vanilla TFLite (built
without QuAKE).

All benchmarking is done with a single CPU thread. Vector instructions
and auto-vectorization are disabled through compiler flags. These
settings are to ensure uniformity in hardware features used by all
operators used in the models. Two warm-up iterations are executed
before each actual benchmark to avoid caching-related slowdowns. CPU
frequency scaling is disabled, and process priority is set to maximum.

\subsubsection{Overall Speedup}\label{subsec:Total-Speedup}

\begin{table*}
\centering
\begin{centering}
\begin{tabular}{ccccccccccccc}
\toprule 
\multirow{2}{*}{Model} & \multicolumn{3}{c}{AMD EPYC} & \multicolumn{3}{c}{Raspberry Pi 5} & \multicolumn{3}{c}{Raspberry Pi Zero 2-W} & \multicolumn{3}{c}{Vision Five 2}\tabularnewline
\cmidrule{2-13}
 & TFLite & QuAKE & QuAKE2 & TFLite & QuAKE & QuAKE2 & TFLite & QuAKE & QuAKE2 & TFLite & QuAKE & QuAKE2\tabularnewline
\midrule 
ViT-Ti & 0.5 & 26.7 & 19.0 & 0.9 & 39.1 & 26.8 & 5.1 & \textbf{23.7} & 15.5 & 21.7 & 3.7 & 2.5\tabularnewline
ViT-B & 7.0 & 12.4 & 10.1 & 11.2 & 22.3 & 16.6 & - & - & - & 252.2 & 2.4 & 1.8\tabularnewline
ViT-L & 16.3 & 8.1 & 6.7 & 24.8 & 16.3 & 12.6 & - & - & - & 553.9 & 1.9 & 1.4\tabularnewline
Swin-Ti & 0.5 & 21.7 & 15.8 & 0.8 & 28.9 & 19.3 & 4.6 & 16.1 & 11.6 & 13.4 & 3.1 & 2.0\tabularnewline
ConvNext-Ti & 0.6 & \textbf{27.5} & \textbf{21.6} & 0.9 & \textbf{46.0} & \textbf{34.3} & 4.9 & 23.4 & \textbf{18.5} & 12.6 & \textbf{5.2} & \textbf{3.9}\tabularnewline
YOLO-v8  & 0.6  & 5.4  & 2.9  & 1.3  & 12.0  & 6.8  & 6.2  & 12.1  & 6.5  & 15.8  & 4.6  & 3.0 \tabularnewline
FastSAM-S & 1.7  & 2.8  & 1.4  & 3.2  & 9.9  & 5.5  & 16.5  & 7.5  & 3.4  & 48.5  & 3.0  & 2.1 \tabularnewline
Whisper-Ti(E) & 1.9 & 19.3 & 12.6 & 3.5 & 28.0 & 18.4 & 19.0 & 22.7 & 14.7 & 130 & 2.5 & 1.7\tabularnewline
Whisper-Ti(D) & 0.02 & 1.8 & 1.9 & 0.04 & 5.6 & 4.0 & 0.3 & 0.8 & 0.8 & 0.4 & 0.4 & 0.2\tabularnewline
Whisper-B(E) & 4.3 & 15.3 & 11.7 & 7.7 & 24.7 & 16.7 & 43.0 & 19.8 & 13.2 & 278.1 & 2.3 & 1.7\tabularnewline
Whisper-B(D) & 0.03 & 1.9 & 1.9 & 0.1 & 4.8 & 10.9 & - & - & - & 0.8 & 0.7 & 0.5\tabularnewline
Whisper-S(E) & 16.3 & 10.9 & 7.5 & 27.7 & 19.2 & 12.9 & - & - & - & 933.2 & 2.1 & 1.4\tabularnewline
Whisper-S(D) & 0.1 & 3.1 & 2.0 & 0.2 & -1.1 & -3.6 & - & - & - & - & - & -\tabularnewline
GPT2-L & 76.7 & 8.0 & 6.1 & 120.8 & 11.3 & 7.9 & - & - & - & 2641.7 & 0.8 & 0.3\tabularnewline
OPT-1.3B & 252.3 & 4.2 & 3.1 & - & - & - & - & - & - & - & - & -\tabularnewline
GPT2-XL & 148.9 & 6.4 & 5.4 & - & - & - & - & - & - & - & - & -\tabularnewline
\midrule 
\textbf{Average} & - & \textbf{11.6} & \textbf{8.8} & - & \textbf{19.7} & \textbf{14.3} & - & \textbf{17.1} & \textbf{12.0} & - & \textbf{2.1} & \textbf{1.4}\tabularnewline
\end{tabular}
\par\end{centering}
\caption{Model inference speedups on each hardware setup. \textquotedblleft TFLite\textquotedblright{}
columns are inference times in seconds without QuAKE, and \textquotedblleft QuAKEx\textquotedblright{}
columns are \% speedups with QuAKEx. The geometric averages of the
speedups are also reported.}
\vspace{-1.5em}
\label{table:all-speedups}
\end{table*}

\noindent We report in Table \ref{table:all-speedups} the inference
times for each model using the vanilla TFLite framework, and the speedups
when using QuAKE and QuAKE2 ops, on each of the hardware setups described
in Section \ref{subsec:Hardware}. Some models could not be run on
all hardware due to memory limitations, and the unavailability of
results for this reason is indicated in Table \ref{table:all-speedups}
by ``-''. We make the following observations along various axes
of the result data.

\paragraph*{Approximation Quality}

QuAKE leads to substantial average speedups of 10\% or greater on
most hardware, topping out at 46\%. QuAKE2 tends to be slower than
QuAKE by about 10\%. Nevertheless, QuAKE2 provides speedup over the
vanilla implementation of 8-14\% on average, reaching a maximum of
34.3\%.

\paragraph*{Hardware}

These speedups are observed in multiple scales of hardware: server,
mobile, and embedded, thus showcasing the hardware-agnostic benefits
of QuAKE. The least speedups are observed with the Vision Five 2,
reaching only 5\% at best. Once cause of these less substantial speedups
is that GEMM is more of a bottleneck on this platform than the others,
typically accounting for more than 95\% of the inference latency.
Overall, speedups are consistent, and follow similar trends across
different classes of hardware.

\paragraph*{Model Characteristics}

As Table \ref{table:all-speedups} shows, speedups remain significant
across the tested size-range for both QuAKE and QuAKE2, with 5\% speedup
for GPT2-XL with 1.5B parameters. Although smaller models tend to
benefit more, the speedups are commensurate with the proportions of
inference time accounted for by each operation listed in Table \ref{table:models},
and consistent with Amdahl's law\cite{comparch-pat-henn}. Transformer
models also tend to benefit more due to Softmax and GELU accounting
for sizable portions of their inference times, usually >10\% combined.
The speedups from QuAKE are by no means restricted to transformers.
In fact, the largest speedups among all models are observed with ConvNeXt
on all hardware, up to 46\% from just GELU activations. YOLO-v8 and
FastSAM-S, which are also CNN models, also exhibit significant speedups
(5-12\% for QuAKE), mainly from the logistic function. These results
exemplify the utility of QuAKE for non-transformer architectures,
especially when using exponential activations in convolutional layers
as with ConvNeXt.

\subsubsection{Speedup from Fusing Affine Input Transformations}\label{subsec:Speedup-affine}

Here, we measure the contributions of fusing affine input transformations,
as discussed in Section \ref{subsec:QuAKE}, to the total inference
speedup. We measure inference speedup for these models over TFLite
operators after disabling these optimizations in QuAKE, and compare
them to the speedups observed when they are enabled. To disable the
optimizations, we do not incorporate the temperature scaling and maximum
subtraction in Softmax, and the constant scale factors in the Tanh-based
GELU approximation. Effectively, we are plugging in the standard exponential
forms of QuAKE into Softmax, and Equation \ref{eq:gelu-quake2} for
GELU. To do so, we create TFLite ``models'' with only a single layer
containing the Softmax or GELU op being benchmarked. The softmax is
computed row-wise on a 16384x16384 input matrix, while for GELU, we
use an input vector with 260k elements. The results are reported in
Table \ref{table:opfusion}. Our proposed optimizations provide up
to 25\% additional speedup for softmax, and 5\% for GELU, with both
QuAKE and QuAKE2. The individual ops themselves see massive speedups
(more than 2x for softmax and 2.5x for GELU) over the vanilla TFLite
ops.

\begin{table}
\centering
\begin{centering}
\begin{tabular}{c|c|c|c|c|c}
\multirow{2}{*}{H/W} & \multirow{2}{*}{Op} & \multicolumn{2}{c|}{%
\begin{minipage}[t]{8em}%
\begin{center}
Speedup w/o Op Fusion (X)
\par\end{center}%
\end{minipage}} & \multicolumn{2}{c}{%
\begin{minipage}[t]{10em}%
\begin{center}
\emph{\uline{Additional}} Speedup due to Op Fusion (X)
\par\end{center}%
\end{minipage}}\tabularnewline
\cline{3-6}
 &  & QuAKE & QuAKE2 & QuAKE & QuAKE2\tabularnewline
\hline 
\multirow{2}{*}{EPYC} & Softmax & 2.41 & 1.73 & 1.18 & 1.17\tabularnewline
\cline{2-2}
 & GELU & 2.85 & 1.73 & 1.08 & 1.05\tabularnewline
\hline 
\multirow{2}{*}{RPi-5} & Softmax & 2.29 & 1.44 & 1.21 & 1.17\tabularnewline
\cline{2-2}
 & GELU & 2.82 & 1.78 & 1.13 & 1.04\tabularnewline
\hline 
\multirow{2}{*}{RPi-0} & Softmax & 2.87 & 1.37 & 1.25 & 1.17\tabularnewline
\cline{2-2}
 & GELU & 1.96 & 1.53 & 1.02 & 1.03\tabularnewline
\hline 
\multirow{2}{*}{VF2} & Softmax & 2.00 & 1.48 & 1.24 & 1.17\tabularnewline
\cline{2-2}
 & GELU & 1.53 & 1.22 & 1.03 & 1.02\tabularnewline
\end{tabular}
\par\end{centering}
\caption{Speedups contributed by fusing affine input transformations}
\vspace{-2em}
\label{table:opfusion}
\end{table}

\subsection{Downstream Task Performance}\label{subsec:Downstream-Task-Performance}

Here, we present the performance results of the models analyzed in
this work on various downstream tasks. For these evaluations, we employ
the same models and inference framework used in Section \ref{subsec:Inference-Speed}.
All the evaluations are performed on the EPYC server.

\subsubsection{Image Classification}

We report classification accuracy (top-1) on the Imagenet\cite{imagenet-data}
(ILSVRC) 2012 validation dataset containing 50,000 images in Table
\ref{table:imagenet-results}. The largest accuracy drop is 0.02\%
with QuAKE, and 0.022\% with QuAKE2. For all models other than ViT-Ti,
QuAKE and QuAKE2 are \emph{more} accurate than the default TFLite
operators.

\begin{table}
\begin{centering}
\begin{tabular}{cccc}
\toprule 
Model & TFLite (\%) & QuAKE ($\Delta$,\%) & QuAKE2 ($\Delta$,\%)\tabularnewline
\midrule 
ViT-Ti & 77.418  & -0.020  & -0.022 \tabularnewline
ViT-B & 84.894  & \textbf{0.024 } & \textbf{0.024 }\tabularnewline
ViT-L & 85.572  & \textbf{0.170 } & \textbf{0.012 }\tabularnewline
Swin-Ti & 80.772  & \textbf{0.020 } & \textbf{0.002 }\tabularnewline
ConvNext-Ti & 81.948  & \textbf{0.002 } & \textbf{0.016 }\tabularnewline
\end{tabular}
\par\end{centering}
\caption{Model accuracy changes with QuAKE on ILSVRC 2012 validation data (more
is better)}
\vspace{-1em}
\label{table:imagenet-results}

\end{table}

\subsubsection{Speech Transcription}\label{subsec:Speech-Transcription}

To evaluate the Whisper models' ASR performance, we use the LibriSpeech\cite{librispeech}
dataset (\href{https://www.openslr.org/12}{test(other) split}). In
this task, the model's decoder predicts the next output token, based
on the encoder output and tokens predicted thus far, in order to obtain
the transcribed text. Performance is measured by the Word Error Rate
(WER) between the ground-truth transcription and the model output.
Table \ref{table:librispeech} compares the performance between QuAKE
and TFLite implementations. On this dataset, QuAKE and QuAKE2 always
lead to an improvement in accuracy. A surprising observation is that
the QuAKE performance is often better than that of QuAKE2 for the
Tiny and Base models, albeit by up to 0.05\%.

\begin{table}
\begin{centering}
\begin{tabular}{cccc}
\toprule 
Model & TFLite (\%) & QuAKE (\%) & QuAKE2 (\%)\tabularnewline
\midrule 
Whisper (Ti) & 19.57 & \textbf{19.53} & \textbf{19.56}\tabularnewline
Whisper (B) & 16.08 & \textbf{15.97} & \textbf{16.02}\tabularnewline
Whisper (S) & 13.53 & \textbf{13.50} & \textbf{13.48}\tabularnewline
\end{tabular}
\par\end{centering}
\caption{Word Error Rates (WER) of Whisper models on LibriSpeech-test(other)
data (lower is better).}
\vspace{-2em}
\label{table:librispeech}
\end{table}

\subsubsection{Natural Language Processing}

We evaluate the GPT and OPT models on the Hellaswag\cite{hellaswag}
benchmark (validation split) with 10,042 multiple-choice reasoning
questions. The results are reported in Table \ref{table:hellaswag}.
Both QuAKE and QuAKE2 yield comparable, albeit marginally higher accuracy,
as observed in Section \ref{subsec:Speech-Transcription}. QuAKE is
often more accurate than QuAKE2 on this dataset as well, albeit by
at most 0.06\%. While a thorough investigation of these counter-intuitive
results is out of the scope of this work, we hypothesize that the
distribution shift between training and validation/test data leads
to many data points being close to a classification boundary, which
happen to get pushed to the correct side due to the perturbations
from QuAKE approximations.

\begin{table}
\begin{centering}
\begin{tabular}{cccc}
\toprule 
Model & TFLite (\%) & QuAKE (\%) & QuAKE2 (\%)\tabularnewline
\midrule 
GPT2-L & 43.14 & \textbf{43.14} & \textbf{43.16}\tabularnewline
OPT-1.3B & 41.95 & \textbf{42.01} & \textbf{41.95}\tabularnewline
GPT2-XL & 48.91 & \textbf{48.97} & \textbf{48.92}\tabularnewline
\end{tabular}
\par\end{centering}
\caption{Hellaswag accuracy results}
\vspace{-2em}
\label{table:hellaswag}

\end{table}

%% file: related-work.tex
\section{Related Work}

Early work used bit-level properties of number representations for
rapidly approximating functions such as the logarithm\cite{mitchell-log},
and the now-famous Fast Inverse Square Root\cite{kahan-frsqrt}\cite{wiki-frsqrt}
algorithm. The use of such an approach for the exponential was proposed
in \cite{schraudolph-exp}. Recently, this approach has been tested
in inference frameworks for edge devices \cite{aifes,kb-iot}. Higher
order extensions, and those that use LUTs to refine the approximation
have also been developed\cite{fexp-combustion}\cite{exp-mantissa-poly}.
However, the speedup from non-linearities and the effects of approximation
on downstream task performance, in the context of model inference,
have not been studied by any of these works. Layer activations other
than the logistic function are not considered, and no enhancements
are made beyond plugging in the approximate exponential function,
such as fusing general affine input transformations into QuAKE. QuAKE2
further differs from the second-order approximation of \cite{exp-mantissa-poly}
in that our choice of coefficients for refining the approximation
yields a simpler approximation that is continuous and accurate for
integers, as described in Section \ref{subsec:QuAKE2}.

There are several other classes of efficient approximations for exponential
non-linearities. A related group of approaches aims to develop efficiently
computable, albeit less accurate approximations such as piecewise
linear functions or ratios of polynomials\cite{ml-plac,peano-vit}.
\cite{twopass-smax} proposes an exponentiation algorithm meant to
produce accurate results with lower-degree polynomials. This is accomplished
using Cody-Waite range reduction\cite{cody-waite-rr}, which rounds
the input to obtain an integer part and a (bounded) fractional part.
QuAKE2 may be viewed as implicitly performing a form of range reduction
by extracting the mantissa of the result, which is subsequently refined
within the range $[0,1)$. LUT-based approaches such as NN-LUT\cite{nn-lut}
can be efficient in custom hardware, but do not scale well as more
accuracy is desired, owing to high memory requirements. Quantization
and low-precision computing has been used to accelerate the computation
of non-linearities, most notably with custom hardware designs\cite{softermax,hw-efficient-softmax}.
Due to its modular and plug-and-play nature, QuAKE can be used orthogonally
or in conjunction with ideas from many of the above works (for instance,
quantization).

%% file: endmatter.tex
\section{Conclusions}

In this work, we developed QuAKE, a novel family of operators that
approximate commonly used exponential non-linearities to accelerate
model inference. We also presented the first known assessment of this
class of approaches over a wide range of model and hardware configurations,
and application domains, with results that strongly affirm their suitability
as drop-in replacements for standard non-linearities. The first-order
QuAKE operators provide the biggest speedups, reaching up to 45\%
while limiting performance degradation to levels that are acceptable
for many applications. The second-order QuAKE2 operators are always
able to match or surpass the performance of standard imlpementations
while still providing substantial speedups of up to 32\%.

Real-time and edge/embedded AI applications such as on-device ASR
and vision are expected to benefit the most from the inference latency
reductions of QuAKE. Our results also demonstrate that the benefits
of QuAKE do extend to large-scale models such as GPT2-XL, and likely
to larger models as well. We also strongly believe that lowering the
computational cost of exponential non-linearities, especially GELU,
will make them viable options for those designing ML models, thus
facilitating the development of better models.

\newpage{}

\input{main.bbl}